\documentclass{article}

\usepackage{microtype}
\usepackage{graphicx}
\usepackage{bbm} 
\usepackage{subcaption}
\usepackage{booktabs} 

\usepackage{hyperref}

\usepackage[preprint]{icml2026}

\usepackage{amsmath}
\usepackage{amssymb}
\usepackage{mathtools}
\usepackage{amsthm}
\usepackage{float}

\usepackage[capitalize,noabbrev]{cleveref}

\theoremstyle{plain}

\theoremstyle{definition}

\theoremstyle{remark}

\usepackage[textsize=tiny]{todonotes}

\icmltitlerunning{Submission and Formatting Instructions for ICML 2026}

\begin{document}

\twocolumn[
  \icmltitle{Detecting Brick Kiln Infrastructure at Scale: Graph, Foundation, and Remote Sensing Models for Satellite Imagery Data}
  %



  \icmlsetsymbol{equal}{*}

  \begin{icmlauthorlist}
    \icmlauthor{Usman Nazir}{University of Oxford}
    \icmlauthor{Xidong Chen}{University of Hong Kong}
    \icmlauthor{Hadia Abu Bakar}{University of Sarghoda}
    \icmlauthor{Hafiz Muhammad Abubakar}{Beaconhouse National University}
    \icmlauthor{Raahim Arbaz}{Beaconhouse National University}
    \icmlauthor{Fezan Rasool}{Lahore University of Management Sciences}
    \icmlauthor{Bin Chen}{University of Hong Kong}
    \icmlauthor{Sara Khalid}{University of Oxford}
  \end{icmlauthorlist}

  \icmlaffiliation{University of Oxford}{Planetary Health Informatics (PHI) Lab, University of Oxford, Oxford, UK.}
  \icmlaffiliation{University of Hong Kong}{Future Urbanity and Sustainable Environment (FUSE) Lab, University of Hong Kong, Hong Kong, China.}
  \icmlaffiliation{Beaconhouse National University}{School of Computer and IT, Beaconhouse National University, Lahore, Pakistan}
  \icmlaffiliation{Lahore University of Management Sciences}{Computer Science Department, Lahore University of Management Sciences, Lahore, Pakistan}
\icmlaffiliation{University of Sarghoda}{Computer Science Department, University of Sarghoda, Sarghoda, Pakistan}
  \icmlcorrespondingauthor{Usman Nazir}{usman.nazir@ndorms.ox.ac.uk}
  \icmlcorrespondingauthor{Sara Khalid}{sara.khalid@ndorms.ox.ac.uk}

  \icmlkeywords{Machine Learning, ICML}

  \vskip 0.3in
]



\printAffiliationsAndNotice{}  

\begin{abstract}
  Brick kilns are a major source of air pollution and forced labor in South Asia, yet large-scale monitoring remains limited by sparse and outdated ground data. We study brick kiln detection at scale using high-resolution satellite imagery and curate a multi-city zoom-20 resolution (0.149~m\,pixel$^{-1}$) dataset comprising over $1.3$ million image tiles across five regions in South and Central Asia. We propose ClimateGraph, a region-adaptive graph-based model that captures spatial and directional structure in kiln layouts, and evaluate it against established graph learning baselines. In parallel, we assess a remote sensing–based detection pipeline and benchmark it against recent foundation models for satellite imagery. Our results highlight complementary strengths across graph, foundation, and remote sensing approaches, providing practical guidance for scalable brick kiln monitoring from satellite imagery.
\end{abstract}

\section{Introduction}
Forced labor remains one of the most pervasive yet least visible human rights violations of the modern era. According to the Global Slavery Index 2023, an estimated $49.6$ million people worldwide are trapped in forced labor~\cite{walk2023global}. Nearly $29.3$ million of these individuals are concentrated within the so-called Brick Kiln Belt of South Asia, a region spanning approximately $1,551,997$~km\textsuperscript{2} across Afghanistan, Pakistan, India, Bangladesh, and Nepal. Brick kilns in this region are widely documented as sites of bonded and exploitative labor, often operating beyond the reach of effective regulatory oversight. In recognition of this crisis, the United Nations Sustainable Development Goal (SDG)~$8.7$ explicitly calls for the eradication of forced labor. Systematically mapping brick kilns at scale is therefore a critical prerequisite for monitoring, accountability, and intervention. However, progress toward this goal is fundamentally constrained by severe data scarcity: existing information on kiln locations is both spatially sparse and temporally outdated, largely due to the prohibitive cost and logistical complexity of manual surveys.

Recent advances in computer vision and remote sensing offer multiple paradigms \cite{reed2023scalemae, he2022masked, cong2022satmae} for addressing large-scale humanitarian challenges, such as forced labor monitoring and satellite imagery analysis. Foundation models pretrained on large and diverse datasets have demonstrated strong transferability under weak or zero-shot supervision, while classical remote sensing methods provide efficient and interpretable alternatives for large-area monitoring. In parallel, graph-based learning offers a complementary approach by explicitly modeling spatial relationships and regional context. Understanding the relative strengths and limitations of these paradigms is critical for reliable and scalable brick kiln detection.


In this work, we make the following contributions:
\begin{itemize}
\item We curate a large-scale, multi-city dataset of high-resolution (Zoom-20, 0.149~m/pixel) satellite imagery for brick kiln detection, covering five regions across South and Central Asia.
\item We propose ClimateGraph, a region-adaptive graph-based model that incorporates spatial context and directional information to effectively capture characteristic kiln layouts, and demonstrate its advantages over standard graph neural network baselines.
\item We conduct a systematic comparison of graph-based learning, foundation models, and classical remote sensing approaches for brick kiln detection under varying supervision regimes.
\end{itemize}

\section{Literature Review}
In the following, we review prior work on graph neural networks,
foundation models, and classical remote sensing methods, which together
form the primary modeling paradigms explored in this study.
\subsection{Graph Neural Networks}
Graph neural networks (GNNs) provide a principled framework for learning over relational and non-Euclidean data by propagating and aggregating information across graph neighborhoods. Early foundational architectures include Graph Convolutional Networks (GCN), which extend spectral convolutions to graph-structured data \cite{kipf2016semi}, and inductive methods such as GraphSAGE, which learn neighborhood aggregation functions that generalize to unseen nodes \cite{hamilton2017inductive}. Attention-based variants, such as Graph Attention Networks (GAT), further improve expressivity by learning adaptive edge weights through self-attention mechanisms \cite{velivckovic2017graph}. Due to their ability to explicitly encode spatial relationships and relational structure, GNNs have been widely adopted for geospatial analysis and environmental modeling, where interactions between nearby entities and directional context are critical \cite{wu2020comprehensive}.

\subsection{Foundation Models}

Foundation models pretrained at scale have reshaped modern machine learning by enabling strong generalization across downstream tasks. In computer vision, vision-language models such as CLIP leverage large-scale contrastive pretraining in image–text pairs to achieve robust zero-shot performance across diverse visual tasks \cite{radford2021learning}, while promptable segmentation models like SAM further illustrate the benefits of large-scale pretraining for open-world generalization \cite{kirillov2023segment}. Despite their strong transferability, the application of foundation models to specialized domains such as high-resolution satellite imagery often requires careful adaptation, as the shift in the domain between natural images and remote sensing data can substantially affect performance \cite{liu2024remoteclip}.

Motivated by the success of foundation models in natural vision tasks, recent work in remote sensing has focused on adapting large-scale self-supervised pretraining to satellite imagery. The Masked Autoencoder (MAE) and Masked Image Modeling (MIM) have emerged as dominant paradigms, allowing representation learning without dense annotations \cite{he2022masked}. Representative models such as SatMAE extend MIM to multispectral and temporal satellite data \cite{cong2022satmae}, while Scale-MAE incorporates scale-aware pretraining to better handle heterogeneous ground resolutions \cite{reed2023scalemae}. Other large-scale efforts, including RingMo and general foundation models for Earth observation, demonstrate that pretraining on massive, multi-sensor satellite corpora substantially improves downstream performance on detection and land-cover tasks \cite{sun2022ringmo, zhang2024geoscience}. These approaches highlight the promise of foundation models for Earth observation, while also underscoring the challenges posed by sensor diversity, resolution variation, and limited labeled data.

\subsection{Remote Sensing Methods}
Classical remote sensing methods have long been used for land-use analysis in satellite imagery by exploiting spectral, temporal, and geometric priors. Spectral index based approaches, thresholding, and rule-based pipelines leverage characteristic reflectance patterns, seasonal dynamics, and spatial morphology to identify industrial structures without reliance on large labeled datasets. In parallel, earlier learning-based methods in remote sensing employed lightweight convolutional neural networks, such as KilnNet \cite{nazir2020kiln} and Tiny Inception-ResNet \cite{nazir2019tiny}, as well as segmentation and detection models that include UNet \cite{ronneberger2015u} and region-based CNNs \cite{ren2016faster, girshick2015fast}. Although these classical remote sensing approaches are computationally efficient and highly interpretable, their performance often degrades in dense urban environments and under substantial visual variability, motivating systematic comparisons with more recent foundation based and graph-based models for robust and scalable large-scale monitoring.


\section{Dataset and Preprocessing}
This study uses high-resolution satellite imagery collected over five urban and peri-urban regions in South and Central Asia where brick kiln activity is prevalent. The dataset was constructed specifically for this work using publicly available satellite imagery and manually curated annotations.

\subsection{Study Regions}
We consider five cities spanning diverse geographic, climatic, and urban contexts: Lahore (Pakistan), Delhi (India), Kathmandu (Nepal), Gazipur (Bangladesh), and Kabul (Afghanistan). Together, these regions cover a substantial portion of the South Asian Brick Kiln Belt and exhibit significant variation in settlement density, land-use patterns, and kiln layouts, enabling evaluation under heterogeneous conditions.

\subsection{Imagery}
Satellite imagery was collected exhaustively within predefined geographic boundaries for each city to ensure broad spatial coverage. The number of image tiles varies by region due to differences in geographic extent and urban density. A summary of the dataset composition by city is provided in Table~\ref{tab:dataset_stats}.

All images were tiled into fixed-size patches of 256$\times$256 pixels. Tiling was performed without overlap, and all tiles preserve a consistent spatial resolution and color depth, ensuring uniform input dimensions across all models used in this study.

\begin{table}[h]
\centering
\caption{Geographic coverage and image counts for the Zoom-20 ($0.149$ m/pixel) satellite imagery dataset used in this study.}
\label{tab:dataset_stats}
\begin{tabular}{lc}
\toprule
\textbf{City} & \textbf{Image Tiles} \\
\midrule
Lahore (Pakistan) & 450{,}045 \\
Delhi (India) & 520{,}800 \\
Kathmandu (Nepal) & 51{,}480 \\
Gazipur (Bangladesh) & 132{,}384 \\
Kabul (Afghanistan) & 159{,}870 \\
\bottomrule
\end{tabular}
\label{tab:dataset}
\end{table}


\section{Experimental Setup}

\subsection{Evaluation Regimes}
We consider two complementary evaluation regimes reflecting the operational characteristics of different modeling paradigms. For graph-based learning, ClimateGraph is trained and evaluated on a single global graph constructed across all five countries, enabling the model to exploit long-range spatial context and cross-region dependencies. Performance is reported at the country level by aggregating predictions over nodes belonging to each region. For image-based models, including foundation models and the classical remote sensing baseline, evaluation is conducted independently for each country using country-specific image tiles. These models operate on local image evidence without access to cross-country spatial context. This protocol reflects typical deployment settings for tile-based detection methods.

\subsection{Evaluation Metrics}
We report macro-averaged F1 score for detection tasks, computed consistently across all models and cities.

\subsection{Baselines}
ClimateGraph is compared against standard graph learning baselines including GCN, GAT, and GraphSAGE using identical graph constructions and node features. In parallel, the classical remote sensing pipeline is evaluated against foundation models, including CLIP-based and multimodal vision–language models, under comparable inference conditions.

\subsection{Compute and Implementation.}
Due to the collaborative nature of this work, experiments were conducted across multiple computing environments, including consumer GPUs (RTX 3060/3070/3090) and data-center GPUs (A100). All models were implemented using consistent software frameworks, fixed random seeds where applicable, and standardized hyperparameters. Differences in hardware affect wall-clock time but do not impact reported performance metrics.


\section{Methods}
In this section, we describe the modeling approaches evaluated in this study. Section 4.1 introduces ClimateGraph, our proposed graph-based model designed to capture spatial and directional structure in brick kiln layouts. Sections 4.2 and 4.3 present foundation model baselines, including a CLIP-based vision–language model and a multimodal large vision–language model (Rex-Omni), which enable weakly supervised and zero-shot detection from satellite imagery. Section 4.4 describes a classical remote sensing–based detection pipeline that exploits spectral, temporal, and geometric priors for efficient large-scale inference. 

\begin{figure}
    \centering
    \includegraphics[width=0.8\linewidth]{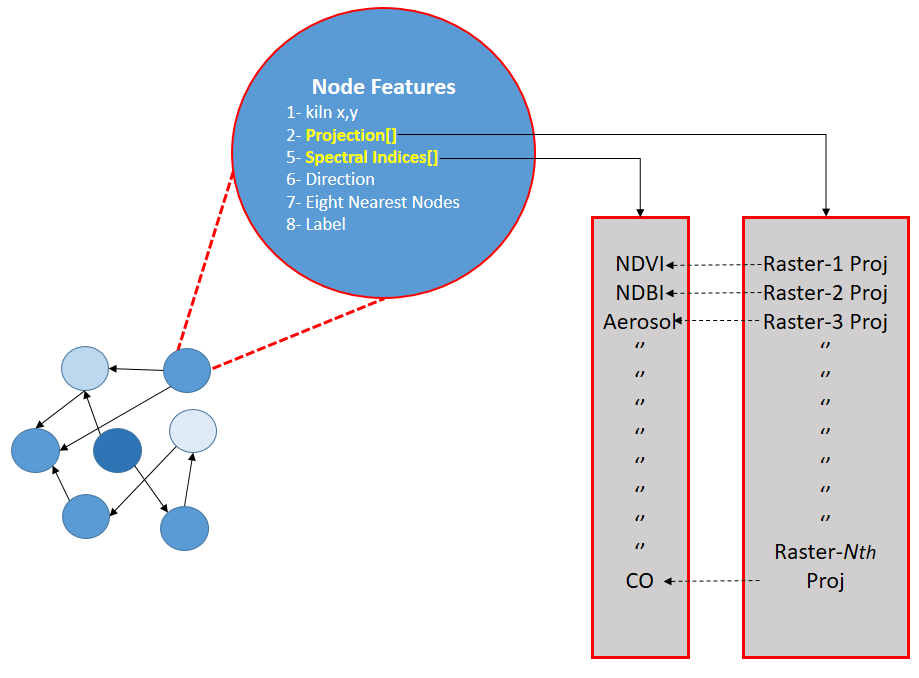}
    \caption{Example of graph construction for ClimateGraph. Nodes represent POIs enriched with multi-modal environmental features from raster datasets. Edges connect each node to its eight nearest neighbors using great-circle distance, with each edge annotated by a bearing angle for directional message passing. This structure enables the model to capture both spatial proximity and orientation, critical for detecting emission sources in heterogeneous landscapes.}
    \label{fig:enter-label}
\end{figure}

\begin{figure*}[h]
    \centering
    \includegraphics[width=0.85\linewidth]{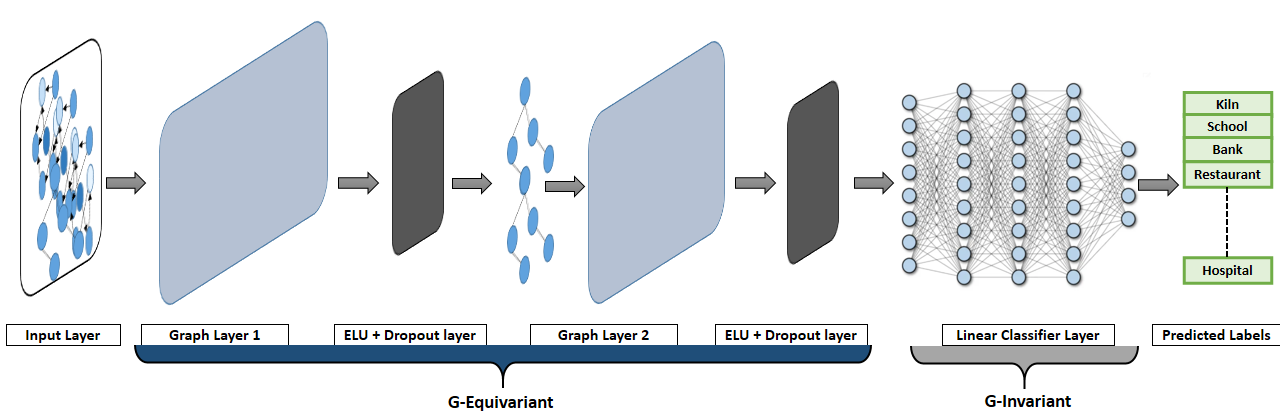}
    \caption{ClimateGraph architecture. The model processes geospatial graphs through stacked anisotropic attention layers that integrate directional kernels with geometry-aware weights. The pipeline transitions from G-equivariant feature extraction, preserving spatial orientation, to G-invariant classification, enabling robust emission source detection across varied geographic contexts.}
    \label{fig:locaGraph}
\end{figure*}
\subsection{ClimateGraph}

Our proposed Region-Adaptive Anisotropic Attention GNN (ClimateGraph) fuses directional kernels with spatially informed attention to combine local geometric precision and broad contextual awareness.
This section details how raw geospatial data are transformed into a graph and
fed to a \emph{Region-Adaptive Anisotropic Attention GNN} (ClimateGraph) for
multiclass prediction.  Points of interest (POIs) are indexed
\(i\in\{1,\dots,N\}\); raster layers are indexed
\(m\in\{1,\dots,M\}\); classes are indexed
\(c\in\{1,\dots,C\}\).


Our pipeline processes geolocated Points of Interest (POIs) with environmental raster features and classifies them into emission source categories. The approach consists of three main components: graph construction, the ClimateGraph layer, and classification.

\subsubsection{Graph Construction}
POIs are connected using $k$-nearest neighbors (great-circle distance). Each edge $(i,j)$ is assigned a bearing angle 
\begin{equation}
    \theta_{ij} = \mathrm{atan2}(\Delta y_{ij}, \Delta x_{ij})
\end{equation}

to capture directionality. Node features are obtained from multi-projection raster sampling, with missing data replaced by buffered mean imputation.

\subsubsection{ClimateGraph Layer}
Let $h_i^{(\ell)} \in \mathbb{R}^{F_\ell}$ be the hidden feature of node $i$ at layer $\ell$. The update rule is:
\begin{equation}
h_i^{(\ell+1)} = \sigma \left( W_0^{(\ell)} h_i^{(\ell)} 
+ \sum_{j \in \mathcal{N}(i)} \alpha_{ij}^{(\ell)} K(\theta_{ij}) W_n^{(\ell)} h_j^{(\ell)} \right),
\end{equation}
where $\sigma$ is a nonlinearity, $W_0^{(\ell)}$ and $W_n^{(\ell)}$ are learnable weight matrices, and $\alpha_{ij}^{(\ell)}$ is a geometry-aware attention score.

The anisotropic kernel $K(\theta)$ is parameterized as a truncated Fourier series:
\begin{equation}
K(\theta) = \sum_{l=0}^{L-1} \kappa_l \cos(l\theta - \mu_l),
\end{equation}
where $\kappa_l$ and $\mu_l$ are learnable amplitudes and phase shifts.

\subsubsection{Geometry-Aware Attention}
Attention scores are computed as:
\begin{equation}
\begin{aligned}
e_{ij}^{(\ell)} = \mathrm{LeakyReLU}\Big(
a^{(\ell)^\top}
\big[
& W_h^{(\ell)} h_i^{(\ell)} \;\| \;
W_h^{(\ell)} h_j^{(\ell)} \\
& \;\| \; \phi(\Delta p_{ij})
\big]
\Big)
\end{aligned}
\end{equation}

\begin{equation}
\alpha_{ij}^{(\ell)} = \frac{\exp(e_{ij}^{(\ell)})}{\sum_{k \in \mathcal{N}(i)} \exp(e_{ik}^{(\ell)})},
\end{equation}
where $\phi(\Delta p_{ij})$ encodes Euclidean distance and sine/cosine of $\theta_{ij}$. 


\subsubsection{Classifier and Loss}
Final node embeddings are passed through a linear classifier:
\begin{equation}
\psi_i = W_c h_i^{(L)} + b_c,
\end{equation}
and trained with a weighted cross-entropy loss to address class imbalance:
\begin{equation}
\mathcal{L} = -\frac{1}{|L|} \sum_{i \in L} w_{y_i} 
\log \frac{\exp(\psi_{i,y_i})}{\sum_{c=1}^C \exp(\psi_{i,c})}.
\end{equation}

\begin{table*}[ht]
\centering
\caption{Performance comparison of graph-based models for brick kiln detection in South Asia. }
\label{tab:model_comparison}
\scalebox{0.9}{\begin{tabular}{lcccc}
\toprule
\textbf{Model} & \textbf{Accuracy (Avg.)} & \textbf{Precision (Avg.)} & \textbf{Recall (Avg.)} & \textbf{F1-score (Avg.)} \\
\midrule
ClimateGraph (Anisotropic) & 0.79 & 0.79 & 0.79 & 0.79 \\
GNN (SAGEConv) \cite{hamilton2017inductive} & 0.78 & 0.78 & 0.78 & 0.78 \\
GAT \cite{velivckovic2017graph}& 0.62 & 0.62 & 0.62 & 0.62 \\
GCN \cite{kipf2016semi}& 0.62 & 0.62 & 0.62 & 0.62 \\
\bottomrule
\end{tabular}}
\end{table*}
\subsubsection{Experimental Setup}
We evaluated on the brick kiln dataset covering the South Asian region (see table~\ref{tab:dataset}). Input features include spectral indices, land cover, and climate variables from multi-projection rasters. Edges are formed using 8-NN with bearing-angle attributes. Experimental evaluations were conducted on a system equipped with 16 GB of RAM, an AMD Ryzen 7 5800HS processor, and an NVIDIA RTX 3060 GPU.

\subsubsection{Quantitative Evaluation}
Baseline models include SAGEConv \cite{hamilton2017inductive}, GAT \cite{velivckovic2017graph}, and GCN \cite{kipf2016semi}. Metrics are macro-averaged Accuracy, Precision, Recall, and F1. ClimateGraph achieves the highest performance across all metrics, with a notable +17\% F1 improvement over isotropic baselines (GAT, GCN) and +1\% over SAGEConv. Gains are especially pronounced in regions with strong spatial anisotropy.

\subsection{CLIP-based Model}

RemoteCLIP is a CLIP-based model that adapts OpenAI's CLIP ViT-B/32 architecture through contrastive fine-tuning on satellite-specific datasets (GeoPairs, LaRS), enabling robust zero-shot image-text matching for remote sensing tasks. The vision encoder extracts 512-dimensional embeddings from 256×256 z20 tiles, capturing spectral and textural features critical for distinguishing industrial structures from natural landcover.

\subsubsection{Multimodal Fusion} 
RemoteCLIP employs dual-encoder architecture with separate ViT-B/32 vision and text transformers. Fusion occurs in shared projection head (512-dim) via contrastive loss, aligning satellite imagery embeddings with geo-specific text descriptions. No cross-attention between modalities during inference.

\subsubsection{Architecture}
\begin{figure}
    \centering
    \includegraphics[width=0.9\linewidth]{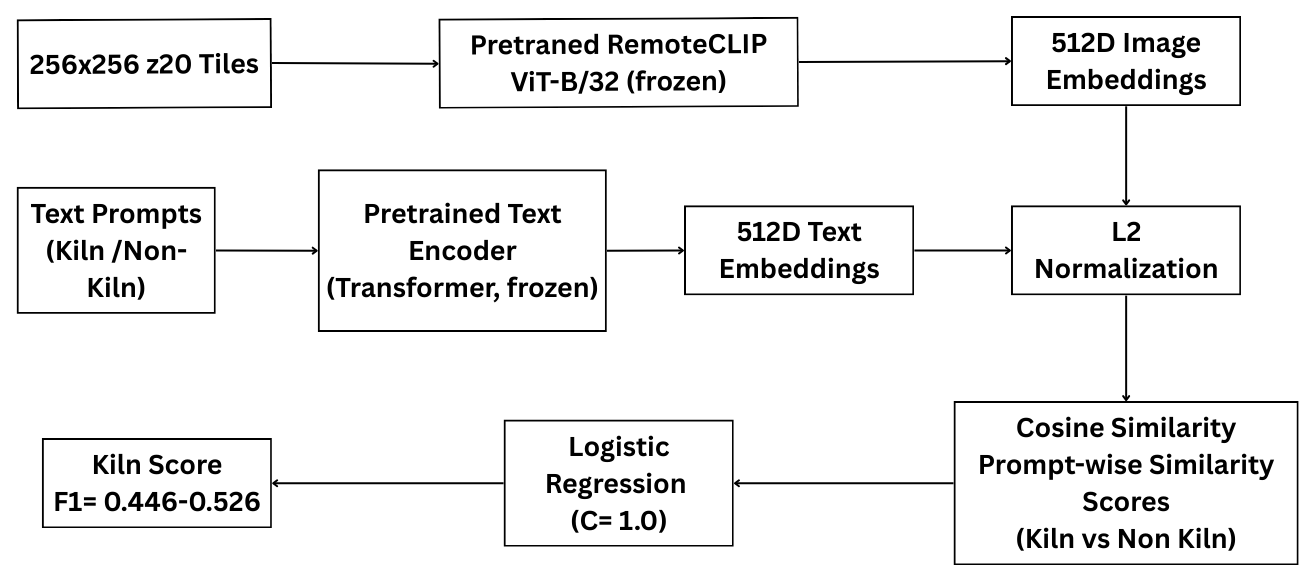}
    \caption{RemoteCLIP zero-shot classification pipeline.}
    \label{fig:remoteclip}
\end{figure}

The vision encoder employs ViT-B/32 with 86M parameters, comprising 12 transformer layers, 768 hidden dimensions, and 12 attention heads. It processes 224×224 inputs via 32×32 patches (64 total), downsampled from 256×256 z20 tiles, using 2D sinusoidal positional embeddings plus CLS token. The text encoder is a Transformer with 63M parameters, supporting 77-token sequences such as "circular brick kiln with chimney stack", projecting to 512-dim L2-normalized embeddings. Training uses asymmetric InfoNCE loss on 2.1M GeoPairs + 1.2M LaRS pairs. All 256×256 z20 tiles are resized to 224×224 to match ViT-B/32 input resolution.
Figure~\ref{fig:remoteclip} illustrates the zero-shot classification pipeline: dual encoders extract L2-normalized 512D embeddings from 256×256 Planet z20 tiles and kiln prompts, computing cosine similarities calibrated via logistic regression (C=1.0).

\subsubsection{Training Procedure}
RemoteCLIP is pretrained in three stages as described in Liu et al. \cite{liu2024remoteclip}. The model first leverages OpenAI CLIP pretraining on 400M web-scale image-text pairs, followed by adaptation on the GeoPairs dataset containing 2.1M satellite-text pairs, and final fine-tuning on the LaRS dataset with 1.2M land remote sensing pairs. This pretraining achieves 85.2\% zero-shot accuracy on LaRS. We use the released pretrained RemoteCLIP ViT-B/32 weights without any additional fine-tuning.
\subsubsection{Contrastive Loss}
We describe the contrastive objective used during RemoteCLIP pretraining for completeness. RemoteCLIP optimizes asymmetric InfoNCE loss. To align vision and text embeddings, we compute:
\begin{equation}
\mathcal{L} = -\sum_{i=1}^N \log \frac{\exp(\text{sim}(v_i,t_i)/\tau)}{\sum_{j=1}^N \exp(\text{sim}(v_i,t_j)/\tau)}, \label{eq:remoteclip}
\end{equation}
where $v_i$, $t_i$ denote vision and text embeddings from the $i$-th tile-prompt pair, $\text{sim}(v,t) = v^\top t/(\|v\|\|t\|)$ is cosine similarity, $\tau=0.07$ is temperature, and $N=4096$ is batch size.

\subsubsection{Text Prompts}
RemoteCLIP uses two engineered prompts for brick kiln discrimination: (1) \textbf{Kiln:} ``Circular brick kiln structure with central chimney stack''; (2) \textbf{Non-kiln:} ``Agricultural fields with crop patterns''.

\subsubsection{Inference}
Inference runs at 0.2s/tile on RTX 3090. Cosine similarity between tile embeddings and prompt embeddings provides a zero-shot semantic score \cite{lee2021scalable}, which we calibrate using logistic regression (C=1.0, stratified 70/30 split on 500 tiles/city). The classifier was trained on 350 tiles and evaluated on 150 held-out tiles per city, yielding F1 scores of 0.446 to 0.526 across five countries without bounding box supervision or domain-specific fine-tuning.

\subsubsection{Quantitative Evaluation}
RemoteCLIP achieves F1 scores of 0.526 (Lahore), 0.465 (Delhi), 0.446 (Kathmandu), 0.491 (Gazipur), and 0.471 (Kabul) across 500 z20 tiles per city. The model demonstrates robust zero-shot transfer from satellite pretraining, with highest performance in Lahore and Gazipur where kiln morphology aligns with training distributions. All evaluations use logistic regression on cosine similarity scores (C=1.0, stratified 70/30 split) without bounding box supervision.

\begin{table*}[h]
\centering
\caption{Performance comparison of foundation models and remote sensing baseline for brick kiln detection across cities in South Asia. 
Each cell reports F1 score.}
\begin{tabular}{lccccc}
\toprule
Model & Lahore & Delhi & Kathmandu & Gazipur & Kabul \\
\midrule
RemoteCLIP & 0.526 & 0.465 & 0.446 & 0.491 & 0.471 \\
Rex-Omni & 0.25 & 0.20 & 0.18 & 0.25 & 0.15 \\
Remote Sensing &  0.533  &  0.305  &  0.166  &  0.650  &  0.436  \\
\bottomrule
\end{tabular}
\label{tab:foundationAndRemoteSensingResults}
\end{table*}

\subsection{Rex-Omni Model}
Rex-Omni leverages Qwen2.5-VL-3B-Instruct (3B parameters) as the multimodal backbone, combining a frozen vision encoder with a transformer decoder trained to autoregressively predict bounding box coordinates as tokenized sequences \cite{mondal2024scalable}. The model processes 256×256 tiles and outputs variable-length coordinate sequences in [x0,y0,x1,y1] format (0-999 normalized bins), enabling zero-shot detection without training data.
\subsubsection{Multimodal Fusion} Qwen2.5-VL-3B fuses vision (ViT-g/14@336px) and language via cross-attention in transformer decoder layers. Visual tokens (576 from 24×24 patch grid) interleave with text tokens, enabling joint reasoning over image content and detection prompts. 

\textbf{Bidirectional fusion:} vision attends to text (prompt grounding) and text attends to vision (spatial localization).
\subsubsection{Architecture}
\begin{figure}[H]
    \centering
    \includegraphics[width=1.0\linewidth]{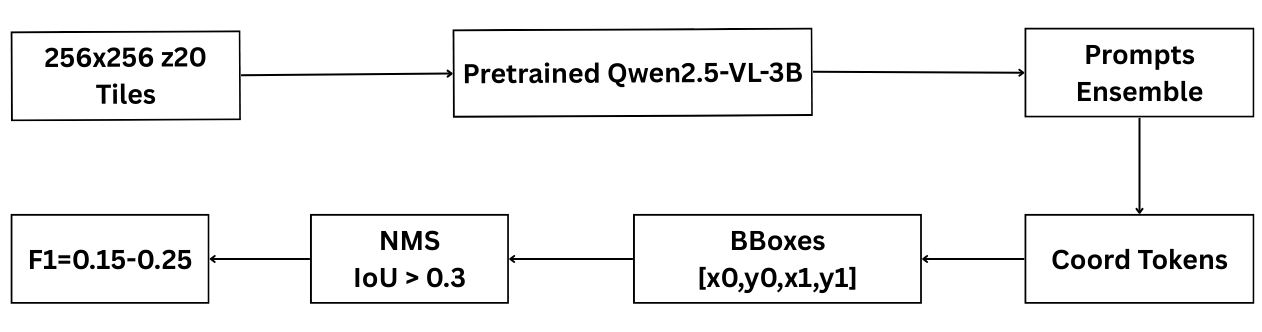}
    \caption{Rex-Omni zero-shot detection pipeline.}
    \label{fig:rexomni}
\end{figure}

Rex-Omni uses Qwen2.5-VL-3B-Instruct (3B parameters) with ViT-g/14 vision backbone (84 layers, 1.6B parameters). All 256×256 z20 tiles are resized to 336px for ViT-g/14 input, generating 576 visual tokens from 24×24 patch grid at 8.2s/256×256 image on A100 (19.4 tokens/sec). Coordinates are quantized into 1,000-bin vocabulary (0-999), yielding 4 tokens per bbox $[x_0,y_0,x_1,y_1]$ predicted sequentially via next-token prediction.
Figure~\ref{fig:rexomni} illustrates the zero-shot detection pipeline: eight-prompt ensemble processed by pretrained Qwen2.5-VL-3B generates coordinate tokens, decoded to bounding boxes, then filtered by NMS (IoU$>$0.3, $\geq$5/8 prompt agreement).

\subsubsection{Training Procedure}
Rex-Omni is pretrained via two stages as described in Mondal et al. \cite{mondal2024scalable}. Stage 1 performs supervised fine-tuning (SFT) on 22M detection samples (COCO, Objects365, custom satellite) using teacher-forced coordinate tokens and L1 bbox regression loss. Stage 2 applies GRPO reinforcement learning with geometry-aware rewards over 50k steps using the reward function:
\begin{equation}
R = \mathbbm{1}(\text{IoU}>0.5) - \lambda\sum_{i\neq j}\mathbbm{1}(\text{IoU}_{ij}>0.7), \label{eq:grpo}
\end{equation}
where $\lambda=0.1$. We use the released pretrained Rex-Omni weights without additional fine-tuning.

\subsubsection{Prompts ensemble for robust detection:}
Rex-Omni uses an 8-prompt ensemble for robust detection: (1) ``Detect circular brick kilns with central chimney''; (2) ``Identify oval kiln structures with tall chimney''; (3) ``Chimneys surrounded by circular ground patterns''; (4) ``Long rectangular kiln structures with chimney stacks''; (5) ``Structures used for firing or baking bricks''; (6) ``Count all brick kilns, especially circular ones''; (7) ``Brick kiln facilities by shape and chimney presence''; (8) ``Industrial footprint of brick manufacturing''.

\subsubsection{Quantitative Evaluation}
\textbf{Dataset:} 2,500 Planet z20 tiles (500/city) from Lahore (PK), Delhi (IN), Kathmandu (NP), Gazipur (BD), Kabul (AF). Rex-Omni yields zero-shot detection success rates of 25\% (Lahore), 20\% (Delhi), 18\% (Kathmandu), 25\% (Gazipur), and 15\% (Kabul), measured as percentage of images with $\geq$1 valid bbox from 8-prompt ensemble. Gazipur exhibits strongest performance due to distinct circular kiln morphology. Detection succeeds when $\geq5/8$ prompts produce overlapping bboxes with $\text{IoU}>0.3$, computed as:
\begin{equation}
\text{IoU}(B_i,B_j) = \frac{|B_i\cap B_j|}{|B_i\cup B_j|}, \label{eq:iou}
\end{equation}
where $B_i = [x_0,y_0,x_1,y_1]$ denotes the $i$-th predicted bounding box, without task-specific fine-tuning.


\subsection{Remote Sensing Baseline}

We adopt a remote-sensing-based baseline to generate brick kiln candidates from satellite imagery. The method is fully deterministic, does not require training data, and is designed for efficient large-scale deployment. It exploits characteristic spectral, temporal, and geometric properties of operational brick kilns through a pixel-wise identification pipeline.

\subsubsection{Baseline Overview:} 
\begin{figure}[H]
\centering
\includegraphics[width=\linewidth]{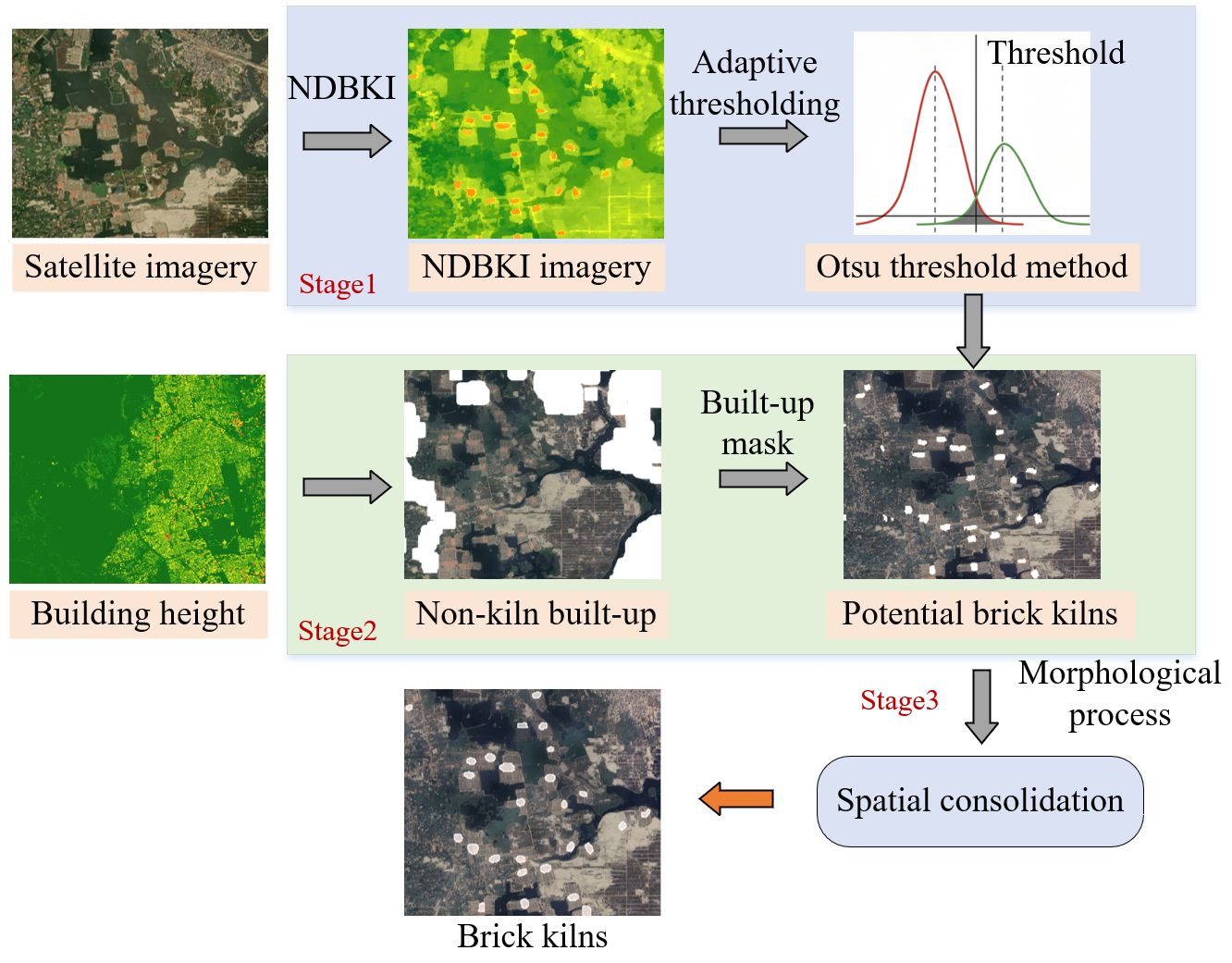}
\caption{Remote sensing baseline pipeline for brick kiln detection.}
\label{fig:rs_baseline_pipeline}
\end{figure}

The baseline operates in three stages as shown in Fig.~\ref{fig:rs_baseline_pipeline}:
(1) spectral--temporal candidate extraction,
(2) built-environment filtering using height priors,
and (3) spatial consolidation.
The output is a set of vectorized brick kiln regions.

\subsubsection{Training Procedure:}

\textbf{Stage 1:} Spectral--temporal candidate extraction

\paragraph{Normalized Difference Brick Kiln Index (NDBKI).} Operational brick kilns typically exhibit a pronounced red appearance
due to exposed fired bricks and combustion residues.
To quantify this property, we compute a Normalized Difference Brick Kiln Index
(NDBKI) from RGB bands as:
\begin{equation}
\mathrm{NDBKI} =
\frac{R - \max(G, B)}{R + \max(G, B)},
\end{equation}
where $R$, $G$, and $B$ denote the red, green, and blue reflectance bands,
respectively.

To account for intermittent kiln operation,
NDBKI is computed for all available observations within a year
and aggregated using an annual 80th-percentile composite,
denoted as NDBKI$_{80}$, which emphasizes periods of active firing
while suppressing background variability:
\begin{equation}
\mathrm{NDBKI}_{80}(x)
=
\operatorname{Percentile}_{80}
\left(
\left\{ \mathrm{NDBKI}_t(x) \right\}_{t=1}^{T}
\right).
\end{equation}

\paragraph{Adaptive Thresholding.}
An adaptive Otsu threshold is applied to the annual NDBKI$_{80}$ composite to separate kiln-like pixels from the surrounding landscape. Pixels exceeding the threshold and corresponding to local maxima in NDBKI$_{80}$ are labeled as potential brick kiln pixels 
(BK$_p$):
\begin{equation}
\mathrm{BK}_p(x) =
\mathbb{I}\big( \mathrm{NDBKI}_{80}(x) > \tau_{\mathrm{Otsu}} \big)
\cdot
\mathbb{I}\big( x \in \mathcal{L}_{\mathrm{max}} \big),
\end{equation}

where $\mathcal{L}_{\mathrm{max}}$ is defined as
\begin{equation}
\mathcal{L}_{\mathrm{max}} =
\left\{
x \;\middle|\;
\mathrm{NDBKI}_{80}(x) \ge \mathrm{NDBKI}_{80}(y),
\;\forall y \in \mathcal{N}_{9\times9}(x)
\right\},
\label{eq:local_max}
\end{equation}

where $\mathcal{N}_{9\times9}(x)$ denotes the $9 \times 9$ pixel neighborhood centered at pixel $x$, corresponding to the typical spatial footprint of brick kilns.

\textbf{Stage 2:} Built-environment filtering

To suppress false positives from red urban buildings,
building height information from the Global 4-m Open Buildings 2.5D dataset
is incorporated as a geometric prior.
A connected region $C$ is classified as non-kiln if
more than 10\% of buildings exceed 3\,m in height:
\begin{equation}
\mathrm{NonKiln}(C)
=
\mathbb{I}
\left(
\frac{1}{|C|}
\sum_{i \in C}
\mathbb{I}(h_i > 3\,\mathrm{m})
> 0.1
\right).
\end{equation}

\textbf{Stage 3:} Spatial consolidation

A morphological closing operation is applied to remove isolated noise
and merge adjacent kiln fragments into spatially contiguous regions:
\begin{equation}
\mathrm{BK}_p^{\ast}
=
(\mathrm{BK}_p \oplus \mathcal{S}) \ominus \mathcal{S},
\end{equation}
where $\oplus$ and $\ominus$ denote dilation and erosion,
and $\mathcal{S}$ is a structuring element. All retained pixels are subsequently
converted into vector polygon representations.

\subsubsection{Inference Speed:}
The baseline operates without any learning or optimization procedures.
Processing a image requires only pixel-wise
index computation, Otsu thresholding, and basic morphological operations,
enabling efficient large-scale inference over national
and continental extents, with approximately 0.5s/256×256 image on a laptop equipped with an NVIDIA RTX 3070 GPU.

\subsubsection{Quantitative Evaluation}

We evaluate the remote sensing baseline for brick kiln detection across five cities: Lahore (Pakistan), Delhi (India), Kathmandu (Nepal), Gazipur (Bangladesh), and Kabul (Afghanistan). Performance is reported using object-level F1 scores, where predicted kiln polygons are matched to ground-truth annotations based on an IoU threshold of 0.3. The baseline achieves higher performance in Gazipur and Lahore, where brick kilns exhibit distinctive and spatially coherent morphologies and are relatively isolated from surrounding built-up areas. In contrast, performance is lower in Kathmandu, where some kilns closely resemble built-up structures and are embedded within
large contiguous urban regions, reflecting increased landscape heterogeneity and fragmented kiln distributions. These results establish the remote sensing baseline as a strong, interpretable, and training-free reference for large-scale brick kiln detection.

\section{Results Discussion}
Across all experiments, our proposed model ClimateGraph, consistently achieves the strongest overall performance when evaluated at scale. By constructing a global graph across all countries, ClimateGraph effectively leverages long-range spatial dependencies, directional relationships, and regional context that are not accessible to tile-based image models. This design enables robust generalization across heterogeneous geographic regions and yields substantial improvements over isotropic graph baselines such as GCN, GAT, and GraphSAGE. 
ClimateGraph achieves the highest scores for precision, precision, recall, and macro-F1, outperforming the isotropic GNN baselines (GAT, GCN) by $17$ percentage points in macro-F1 and surpassing SAGEConv by $1$ percentage points. Metrics are averaged over multiple experimental runs.

The classical remote sensing baseline provides a complementary perspective. Despite lacking learned representations, it offers efficient, interpretable, and scalable detection based on spectral, temporal, and geometric priors. The classical remote sensing baseline provides a complementary perspective. Despite lacking learned representations, it offers efficient, interpretable, and scalable detection based on spectral, temporal, and geometric priors. By relying on deterministic rules rather than data-driven optimization, the baseline remains robust under limited or heterogeneous supervision and enables consistent large-area deployment. While its performance is generally lower than graph-based learning and foundation models in complex urban environments, it establishes a strong training-free reference and a meaningful lower bound for brick kiln detection.

In contrast, foundation models show clear promise under weak supervision and zero-shot use, but their performance is more variable across cities than graph-based or classical methods: a CLIP-based model (RemoteCLIP) attains F1 scores of $0.526$ (Lahore), $0.465$ (Delhi), $0.446$ (Kathmandu), $0.491$ (Gazipur), and 0.471 (Kabul); Rex-Omni’s zero-shot coordinate predictions are much lower ($\equiv 0.25–0.15$ F1 across the same cities); the deterministic remote-sensing pipeline nevertheless matches or exceeds CLIP in some regions (e.g., $0.533$ in Lahore and $0.650$ in Gazipur). These numbers are reported in Table~\ref{tab:foundationAndRemoteSensingResults}.s




\section{Conclusion}
We presented a comprehensive study of brick kiln detection at scale using high-resolution satellite imagery across South and Central Asia. We introduced ClimateGraph, a region-adaptive graph-based model that effectively captures spatial and directional structure, and demonstrated its advantages over standard graph baselines. In parallel, we benchmarked classical remote sensing methods against modern foundation models, revealing complementary strengths across modeling paradigms under varying supervision regimes. By unifying graph-based, foundation, and remote sensing approaches within a common evaluation framework, our work provides practical guidance for scalable and data-efficient monitoring of brick kilns, with broader implications for environmental and humanitarian applications of machine learning.


\section*{Impact Statement}
This work aims to support large-scale monitoring of unregulated brick kilns in South Asia, which are associated with air pollution and forced labor. By enabling scalable detection from satellite imagery, the proposed methods may assist researchers, policymakers, and regulatory agencies in improving situational awareness and prioritizing field interventions. 


\nocite{langley00}

\bibliography{paper}

@inproceedings{girshick2015fast,
  title={Fast r-cnn},
  author={Girshick, Ross},
  booktitle={Proceedings of the IEEE international conference on computer vision},
  pages={1440--1448},
  year={2015}
}

@article{ren2016faster,
  title={Faster R-CNN: Towards real-time object detection with region proposal networks},
  author={Ren, Shaoqing and He, Kaiming and Girshick, Ross and Sun, Jian},
  journal={IEEE transactions on pattern analysis and machine intelligence},
  volume={39},
  number={6},
  pages={1137--1149},
  year={2016},
  publisher={IEEE}
}

@inproceedings{ronneberger2015u,
  title={U-net: Convolutional networks for biomedical image segmentation},
  author={Ronneberger, Olaf and Fischer, Philipp and Brox, Thomas},
  booktitle={International Conference on Medical image computing and computer-assisted intervention},
  pages={234--241},
  year={2015},
  organization={Springer}
}

@inproceedings{nazir2019tiny,
  title={Tiny-Inception-ResNet-v2: Using deep learning for eliminating bonded labors of brick kilns in South Asia},
  author={Nazir, Usman and Khurshid, Numan and Ahmed Bhimra, Muhammad and Taj, Murtaza},
  booktitle={Proceedings of the IEEE/CVF Conference on Computer Vision and Pattern Recognition Workshops},
  pages={39--43},
  year={2019}
}

@article{nazir2020kiln,
  title={Kiln-net: A gated neural network for detection of brick kilns in South Asia},
  author={Nazir, Usman and Mian, Usman Khalid and Sohail, Muhammad Usman and Taj, Murtaza and Uppal, Momin},
  journal={IEEE Journal of Selected Topics in Applied Earth Observations and Remote Sensing},
  volume={13},
  pages={3251--3262},
  year={2020},
  publisher={IEEE}
}

@article{wu2020comprehensive,
  title={A comprehensive survey on graph neural networks},
  author={Wu, Zonghan and Pan, Shirui and Chen, Fengwen and Long, Guodong and Zhang, Chengqi and Yu, Philip S},
  journal={IEEE transactions on neural networks and learning systems},
  volume={32},
  number={1},
  pages={4--24},
  year={2020},
  publisher={IEEE}
}

@article{velivckovic2017graph,
  title={Graph attention networks},
  author={Veli{\v{c}}kovi{\'c}, Petar and Cucurull, Guillem and Casanova, Arantxa and Romero, Adriana and Lio, Pietro and Bengio, Yoshua},
  journal={arXiv preprint arXiv:1710.10903},
  year={2017}
}

@article{hamilton2017inductive,
  title={Inductive representation learning on large graphs},
  author={Hamilton, Will and Ying, Zhitao and Leskovec, Jure},
  journal={Advances in neural information processing systems},
  volume={30},
  year={2017}
}

@article{kipf2016semi,
  title={Semi-supervised classification with graph convolutional networks},
  author={Kipf, TN},
  journal={arXiv preprint arXiv:1609.02907},
  year={2016}
}

@article{walk2023global,
  title={The global slavery index 2023},
  author={Walk, Free},
  year={2023},
  publisher={Walk Free}
}

@article{mondal2024scalable,
  title={Scalable methods for brick kiln detection and compliance monitoring from satellite imagery: A deployment case study in india},
  author={Mondal, Rishabh and Patel, Zeel B and Jani, Vannsh and Batra, Nipun},
  journal={arXiv preprint arXiv:2402.13796},
  year={2024}
}

@article{liu2024remoteclip,
  title={Remoteclip: A vision language foundation model for remote sensing},
  author={Liu, Fan and Chen, Delong and Guan, Zhangqingyun and Zhou, Xiaocong and Zhu, Jiale and Ye, Qiaolin and Fu, Liyong and Zhou, Jun},
  journal={IEEE Transactions on Geoscience and Remote Sensing},
  volume={62},
  pages={1--16},
  year={2024},
  publisher={IEEE}
}

@article{lee2021scalable,
  title={Scalable deep learning to identify brick kilns and aid regulatory capacity},
  author={Lee, Jihyeon and Brooks, Nina R and Tajwar, Fahim and Burke, Marshall and Ermon, Stefano and Lobell, David B and Biswas, Debashish and Luby, Stephen P},
  journal={Proceedings of the National Academy of Sciences},
  volume={118},
  number={17},
  pages={e2018863118},
  year={2021},
  publisher={National Academy of Sciences}
}

@inproceedings{langley00,
 author    = {P. Langley},
 title     = {Crafting Papers on Machine Learning},
 year      = {2000},
 pages     = {1207--1216},
 editor    = {Pat Langley},
 booktitle     = {Proceedings of the 17th International Conference
              on Machine Learning (ICML 2000)},
 address   = {Stanford, CA},
 publisher = {Morgan Kaufmann}
}

@inproceedings{he2022masked,
  title={Masked autoencoders are scalable vision learners},
  author={He, Kaiming and Chen, Xinlei and Xie, Saining and Li, Yanghao and Doll{\'a}r, Piotr and Girshick, Ross},
  booktitle={Proceedings of the IEEE/CVF conference on computer vision and pattern recognition},
  pages={16000--16009},
  year={2022}
}

@inproceedings{kirillov2023segment,
  title={Segment anything},
  author={Kirillov, Alexander and Mintun, Eric and Ravi, Nikhila and Mao, Hanzi and Rolland, Chloe and Gustafson, Laura and Xiao, Tete and Whitehead, Spencer and Berg, Alexander C and Lo, Wan-Yen and others},
  booktitle={Proceedings of the IEEE/CVF international conference on computer vision},
  pages={4015--4026},
  year={2023}
}

@inproceedings{radford2021learning,
  title={Learning transferable visual models from natural language supervision},
  author={Radford, Alec and Kim, Jong Wook and Hallacy, Chris and Ramesh, Aditya and Goh, Gabriel and Agarwal, Sandhini and Sastry, Girish and Askell, Amanda and Mishkin, Pamela and Clark, Jack and others},
  booktitle={International conference on machine learning (ICML)},
  pages={8748--8763},
  year={2021},
  organization={PmLR}
}

@article{cong2022satmae,
  title={SatMAE: Pre-training transformers for temporal and multi-spectral satellite imagery},
  author={Cong, Yezhen and Khanna, Samar and Meng, Chenlin and Liu, Patrick and Rozi, Erik and He, Yutong and Burke, Marshall and Lobell, David and Ermon, Stefano},
  journal={Advances in Neural Information Processing Systems},
  volume={35},
  pages={197--211},
  year={2022}
}

@inproceedings{reed2023scalemae,
  title={Scaling Masked Autoencoders for Remote Sensing},
  author={Reed, Colin and others},
  booktitle={ICCV},
  year={2023}
}

@article{sun2022ringmo,
  title={RingMo: A remote sensing foundation model with masked image modeling},
  author={Sun, Xian and Wang, Peijin and Lu, Wanxuan and Zhu, Zicong and Lu, Xiaonan and He, Qibin and Li, Junxi and Rong, Xuee and Yang, Zhujun and Chang, Hao and others},
  journal={IEEE Transactions on Geoscience and Remote Sensing},
  volume={61},
  pages={1--22},
  year={2022},
  publisher={IEEE}
}

@article{zhang2024geoscience,
  title={When Geoscience Meets Foundation Models: Toward a general geoscience artificial intelligence system},
  author={Zhang, Hao and Xu, Jin-Jian and Cui, Hong-Wei and Li, Lin and Yang, Yaowen and Tang, Chao-Sheng and Boers, Niklas},
  journal={IEEE geoscience and remote sensing magazine},
  year={2024},
  publisher={IEEE}
}
\bibliographystyle{icml2026}




\end{document}